%% file: main.tex
\documentclass[]{spie}  %>>> use for US letter paper
 \usepackage{booktabs}    % For \toprule, \midrule, \bottomrule
\usepackage{multirow}    % For multirow cells
\usepackage{graphicx}    % For figures if needed elsewhere
\usepackage{amsmath}     % For math symbols like \delta

\usepackage{amsmath,amsfonts,amssymb}
\usepackage{amsfonts}  % or \usepackage{amssymb}
\usepackage{graphicx}
\usepackage[colorlinks=true, allcolors=blue]{hyperref}
\usepackage{textcomp}
\usepackage{multirow}
\usepackage{xcolor}
\usepackage{xcolor}
\usepackage[table]{xcolor}
\usepackage[dvipsnames]{xcolor}

\title{Monocular absolute depth estimation from endoscopy via domain-invariant feature learning and latent consistency}

\author[]{Hao Li}
\author[]{Daiwei Lu}
\author[]{Jesse d'Almeida}
\author[]{Dilara Isik}
\author[]{Ehsan Khodapanah Aghdam}
\author[]{Nick DiSanto}
\author[]{Ayberk Acar}
\author[]{Susheela Sharma}
\author[]{Jie Ying Wu}
\author[]{Robert J.\ Webster III}
\author[]{Ipek Oguz}

\affil[]{Vanderbilt University}

\begin{document}

\maketitle

\begin{abstract}
Monocular depth estimation (MDE) is a critical task to guide autonomous medical robots. However, obtaining absolute (metric) depth from an endoscopy camera in surgical scenes is difficult, which limits supervised learning of depth on real endoscopic images. Current image-level unsupervised domain adaptation methods translate synthetic images with known depth maps into the style of real endoscopic frames and train depth networks using these translated images with their corresponding depth maps. However a domain gap often remains between real and translated synthetic images. In this paper, we present a latent feature alignment method to improve absolute depth estimation by reducing this domain gap in the context of endoscopic videos of the central airway. Our methods are agnostic to the image translation process and focus on the depth estimation itself. Specifically, the depth network takes translated synthetic and real endoscopic frames as input and learns latent domain-invariant features via adversarial learning and directional feature consistency. The evaluation is conducted on endoscopic videos of central airway phantoms with manually aligned absolute depth maps. Compared to state-of-the-art MDE methods, our approach achieves superior performance on both absolute and relative depth metrics, and consistently improves results across various backbones and pretrained weights. Our code is available at \url{https://github.com/MedICL-VU/MDE}.

\end{abstract}

% \keywords{Metric depth map, central airway obstruction, autonomous surgical robot}

\input{_1_introduction}

\input{_2_methods}

\input{_3_results}

\clearpage
\acknowledgments % equivalent to \section*{ACKNOWLEDGMENTS}  

Research reported in this publication was supported by the Advanced Research Projects Agency for Health (ARPA-H) under Award Number D24AC00415-00. The ARPA-H award provided 90\% of total costs with an award total of up to \$11,935,038. The content is solely the responsibility of the authors and does not necessarily represent the official views of ARPA-H. This work was also supported by the Vanderbilt Institute for Surgery and Engineering (VISE) Seed Grant.

% References
\bibliography{report} % bibliography data in report.bib
\bibliographystyle{spiebib} % makes bibtex use spiebib.bst

\end{document}

%% file: _1_introduction.tex
\section{introduction}
Monocular depth estimation (MDE) predicts a pixel-wise depth map from a single RGB image without requiring data from depth sensors. This task is important in medical surgical robotics, where accurate absolute (metric) depth maps are critical for downstream applications such as localization and 3D reconstruction. Recently, foundation models have shown impressive performance in MDE \cite{yang2024depth,tian2024endoomni,piccinelli2025unidepthv2,liang2025distilling}. However, most of these methods focus on relative depth estimation, where the absolute scale of the predicted depth map remains unknown. Due to differences in scale, many of these models require finetuning to generalize to unseen domains, particularly for medical applications where domain shifts can be significant and the model may fail to produce accurate depth maps \cite{han2024depth}. For medical endoscopic images, obtaining ground truth absolute depth maps for finetuning is challenging, primarily due to the incompatibility of depth sensors with the size constraints of endoscopic tools.

% To address this limitation, the common solution is to generate paired \xxx{synthetic} data for supervision. A prior study proposed aligning point clouds from pre-operative CT scans with Structure-from-Motion (SfM) \cite{schonberger2016structure} reconstructions from endoscopic videos to derive such paired depth data \cite{lu2025kidney}. However, this approach may not be suitable for central airway surgical scenes \cite{li2025automated} because the smooth low-texture surfaces lack sufficient features \cite{barbed2023tracking} for SfM to reconstruct a reliable point cloud.

\begin{figure}[!t]
\centering
\includegraphics[width=0.8\linewidth]{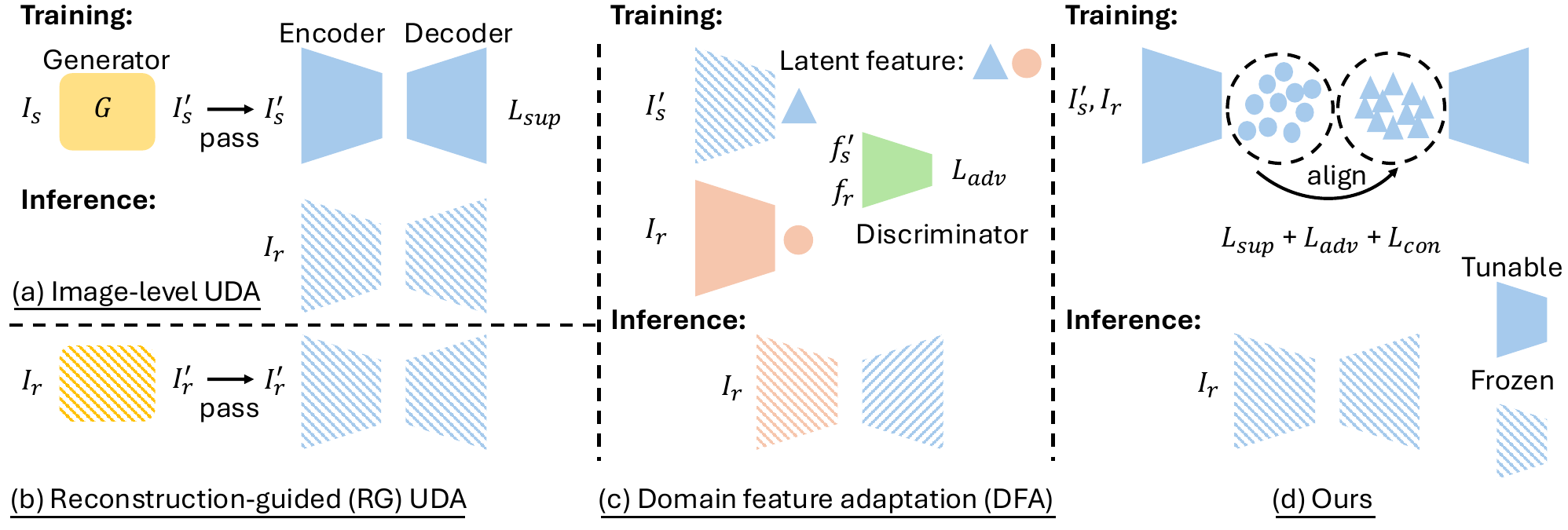}
\caption{Current and proposed methods for monocular depth estimation. (a) \textbf{UDA}: Synthetic image ($I_s$) is first translated into the style of real endoscopic images to obtain $I_s'$, which are then used to train a depth network. During inference, real image ($I_r$) is directly passed into the pretrained (fixed) depth network. (b) \textbf{RG-UDA}: As a close variant, $I_r$ is first passed through the generator to align with the generative training distribution before depth estimation.
(c) \textbf{DFA}: With the pretrained depth network from (a), a separate encoder is trained for $I_r$ using a discriminator and adversarial loss ($L_{\text{adv}}$) to reduce the domain gap. At inference, the encoder and the pretrained depth decoder are used to predict depth from $I_r$.
(d) \textbf{Ours}: Features from the ${I_s}'$ and $I_r$ are aligned in the latent space during training using supervised ($L_{\text{sup}}$), adversarial ($L_{\text{adv}}$), and consistency ($L_{\text{con}}$) losses. Unlike (a–c), which only partially leverage domain information, our approach updates the entire depth network using both domains during training to provide better adaptation to $I_r$.}
\label{framework}

\end{figure}

A common solution leverages unpaired synthetic images with known absolute depth maps % to predict depth from real endoscopic images, 
using unpaired domain adaptation (UDA) \cite{widya2021self,banach2021visually,wang2024structure,xiong2025pps,rau2023task,karaoglu2021adversarial} (Fig.~\ref{framework}(a-b)). These methods involve translating synthetic images into the style of real endoscopic frames with image-to-image translation networks \cite{zhu2017unpaired,park2020contrastive} to reduce the domain gap. Then, a depth network is trained on the translated images using their corresponding absolute depth maps as supervision. Similarly, reconstruction-guided (RG) UDA inputs real images into a trained image translator before the depth network to further reduce the domain gap. In contrast, domain feature adaptation (DFA)
%domain adversarial learning
(Fig.~\ref{framework}(c)) improves performance by training a separate encoder on real endoscopic images; a discriminator guides this encoder to produce feature representations that align with those from a pretrained encoder on synthetic paired data \cite{karaoglu2021adversarial}.

However, both methods face challenges. For %image-level translation 
image-level UDA methods, even with high quality  translated images \cite{wang2024structure,xiong2025pps}, the depth network may still suffer from a residual domain gap between translated synthetic images and real endoscopic images. This gap can lead to biased depth estimation performance in practical applications, even within the same modality \cite{tian2024endoomni}. In the DFA setting \cite{karaoglu2021adversarial}, the decoder remains fixed after being trained only on synthetic features, making it sensitive to misaligned representations from real data.

In this paper, we propose an unsupervised latent space feature alignment method to reduce the domain gap between unpaired translated synthetic and real endoscopic images for accurate absolute depth estimation (Fig.~\ref{framework}(d)). Since existing image translation methods already produce visually realistic outputs, we shift our focus to aligning the latent representations of translated synthetic and real images. We train a discriminator to let the encoder produce domain-invariant feature representations, and apply a consistency regularization to minimize the difference between the latent representations of the two domains. We jointly optimize the entire network on both real and translated synthetic images to produce reliable and accurate depth maps for real central airway obstruction (CAO) endoscopic scenes. Compared to state-of-the-art methods, our approach achieves superior performance on both absolute and relative metrics, which is also observed across different backbone and pretrained weights.
% \xx{Moreover, we evaluate its usability with existing depth estimation models using a variety of pretrained weights and backbones, and observe consistent performance improvements.} 
% Our approach consistently improves results across different pretrained weights and backbones, which indicates that it can be used as a plug-and-play module for existing approaches.

%% file: _2_methods.tex
\section{Materials and Methods}

%\noindent \textbf{\underline{Datasets.}} We extract 4895 synthetic frames with ground truth depths from 11 CAO phantom CTs \cite{li2025automated}, and 2193 real frames from the CAO videos. These frames are used as the training set for both the image translation and depth estimation networks. The test set consists of 90 real CAO frames from three completely held-out CAO phantoms. Each frame is manually aligned to its corresponding CT segmentation to obtain the absolute depth map. Due to potential misalignment in this registration process, two types of masks are applied during evaluation to ensure robustness (see Fig.~\ref{backbone}(c-d)).
\noindent \textbf{\underline{Datasets.}} We create 4,895 synthetic frames with paired ground truth depths from 11 physical CAO phantoms \cite{li2025automated} by acquiring a CT scan of each phantom, manually segmenting the CT, and rendering endoscopic frames from simulated camera positions along the medial axis of the segmentation in Unity (\url{unity.com})  \cite{lu2025kidney}. We also extract 2,193 frames from eight real videos of 5 CAO phantoms, which are a subset of the 11 phantoms above. These two sets of frames are used as the training set (in unpaired manner) for both the image translation and depth estimation networks. The test set consists of 90 real CAO frames from three completely unseen CAO phantoms. Each frame is manually aligned to its corresponding CT segmentation to obtain the absolute depth map. Any potential residual misalignment is accounted for in our evaluation scheme (Sec.\ \ref{sec:results}).
All frames are cropped to $1080 \times 1080$ pixels, then resized to $518 \times 518$. ImageNet normalization statistics are used \cite{yang2024depth}.

\noindent \textbf{\underline{Unsupervised domain adaptation framework.}}
Our approach follows the UDA framework (Fig.~\ref{framework}(a)), which is widely used \cite{widya2021self,banach2021visually,wang2024structure,xiong2025pps,rau2023task} for MDE from endoscopic data, and consists of upstream image translation and downstream depth prediction. Specifically, the image generator ($G$) translates synthetic images (${I_s}$) from the synthetic domain ($\mathcal{S}$) into the style of the real domain ($\mathcal{R}$). The depth network is then trained in supervised manner using the translated synthetic images, ${I_s'}$, and their depth maps ($D_s$), and predicts depth for real frames ${I_r}$.

% Following prior work \cite{lu2024assist}, we adopt CUT \cite{park2020contrastive} as $G$ for image-to-image translation. Since CUT produces visually reliable results, we focus on the depth estimation task rather than refining the translation process.

\noindent \textbf{\underline{Real-synthetic joint training.}} We design our training strategy to allow the depth network to learn directly from real images (${I_r}$) during training, rather than relying solely on translated synthetic images ($I_s'$). In contrast to using a separate encoder for DFA (Fig.~\ref{framework}(c)), which limits semantic latent feature representation, our method jointly trains the entire depth network using both $I_s'$ and ${I_r}$ as input.  $I_s'$ are paired with their absolute depths ($D_s$) for supervision, while ${I_r}$ are used to guide the encoder through domain-invariant feature learning and latent consistency, as illustrated in Fig.~\ref{framework}(d). This leads the encoder to extract features that are both semantically meaningful and domain-agnostic, with the goal of improving generalization to real endoscopic scenes.

\noindent \textbf{\underline{Domain-invariant feature learning.}}  
For learning domain-invariant features, we use domain adversarial learning \cite{ganin2016domain} to align the feature distributions of unpaired $I_s'$ and ${I_r}$. Let $f_s$ and $f_r$ denote the latent representations extracted by the shared encoder ($E$) from $I_s'$ and ${I_r}$, respectively. A discriminator (${Dis}$) is trained to classify whether a given feature comes from domain $\mathcal{S}$ or $\mathcal{R}$ ($L_{adv}^{Dis}$), while the encoder is optimized to confuse the discriminator ($L_{adv}^{E}$). This aims to produce domain-invariant representations in a shared latent space, which helps the supervision from $I_s'$
%\xx{translated} synthetic images 
to be transferred to 
%real images
${I_r}$, even without ground truth for the ${I_r}$.

\noindent \textbf{\underline{Latent consistency.}}  
While adversarial learning aligns feature distributions globally, it does not guarantee semantic consistency between real and synthetic latent representations. To address this, we introduce a latent consistency regularization that enforces directional alignment between the features $f_s$ and $f_r$ extracted from $I_s'$ and ${I_r}$. A cosine similarity loss ${L}_{\text{con}}$ is applied during training to maximize agreement between these features. This regularization complements the adversarial objective by minimizing in-domain specific variation and producing more reliable domain-invariant latent features.
This helps preserve semantic structure and improves generalization between domains, as directional alignment more precisely captures semantic similarity.

\noindent \textbf{\underline{Objective function.}}  
Our training objective combines three loss terms: a supervised loss ($L_{{sup}}$), an adversarial loss ($L_{{adv}}$), and a consistency loss ($L_{{con}}$). The objective function is defined as:
\[
L_{{total}} =
L_{{sup}}(\tilde{D_s}, D_s) +
L_{{adv}}^{E}\Big(Dis(f_{r}), \mathcal{S}\Big) + 
\frac{1}{2} \bigg(
L_{{adv}}^{Dis}\Big(Dis(f_s'), \mathcal{S}\Big) +
L_{{adv}}^{Dis}\Big(Dis(f_r), \mathcal{R}\Big)
\bigg) + 
L_{{con}}(f_r, f_s')
\]
where $\tilde{D}_s$ is the prediction from $I_s'$, and $L_{sup}$ is the scale-invariant logarithmic loss \cite{yang2024depth}. The adversarial loss is binary cross-entropy, where the domain labels $\mathcal{S}$ and $\mathcal{R}$ are encoded as 0 and 1. $L_{con}$ is defined as cosine similarity.

%% file: _3_results.tex
\section{Experiments and conclusion}
\label{sec:results}

\noindent \textbf{\underline{Compared methods and implementation details.}} We compare: (1) Depth Anything v2, a foundation model, in zero-shot manner \cite{yang2024depth}, (2) an image-level UDA model \cite{widya2021self}, (3) an RG-UDA model \cite{rau2023task}, and (4) a DFA model \cite{karaoglu2021adversarial}. 
% We avoid end-to-end UDA to prevent translation bias from the generator. As we aim to reduce the domain gap within the depth network, since even well-translated images may still have a domain gap \cite{wang2024structure}.
%Following our prior work \cite{lu2024assist}, w
We use the CUT model \cite{park2020contrastive} as image generator; the depth network is adopted from Depth Anything v2 (ViTs) \cite{yang2024depth}. 
 %The depth networks are initialized with pretrained weights from two  depth foundation models \cite{yang2024depth,tian2024endoomni}. 
Following prior work \cite{yang2024depth,karaoglu2021adversarial}, we set the number of epochs to 30 and the initial learning rate to $5e^{-6}$ with a polynomial decay (power 0.9) applied at each epoch. 
% The learning rate decays progressively at each epoch using a polynomial schedule with power 0.9.
The AdamW optimizer is used, and we used an NVIDIA A6000.

\input{table/main_table}

\begin{figure}[t]
\centering
\includegraphics[width=0.9\linewidth]{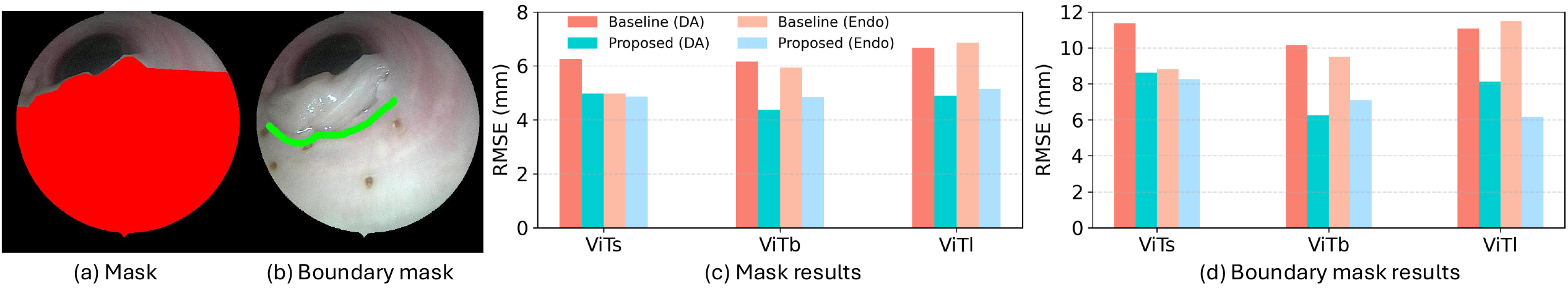}
\caption{(a-b) Illustrations of masks.
(c-d) RMSE comparison between different pretrained weights and backbone sizes. ``DA'' and ``Endo'' denote the pretrained weights from Depth Anything v2 (Metric) \cite{yang2024depth} and EndoOmni \cite{tian2024endoomni}, respectively.}
\label{backbone}
\end{figure}

\noindent \textbf{\underline{Evaluation metrics.}} Manual alignment for the ground truth depth map is prone to residual errors in the 2D-to-3D mapping. Since the manual alignment focuses on the tumor bottom boundary, we manually create two ROI masks for depth estimation evaluation (Fig.~\ref{backbone}(a–b)). The ``mask'' measures overall accuracy to avoid regions behind the tumor that are prone to registration error. The ``boundary mask'' only focuses on the proximal tumor boundary, which is most important for potential downstream tasks such as tumor resection \cite{acar2025monocular}. Given a mask $M$, we report threshold accuracy $\delta_1$, defined as the percentage of pixels $i \in M$ where $\max\left(\frac{\tilde{D}_i}{D_i}, \frac{D_i}{\tilde{D}_i}\right)<1.25$; Absolute Relative Error (AbsRel) $= \frac{1}{|M|} \sum_{i \in M} \frac{|\tilde{D}_i - D_i|}{D_i}$; and Root Mean Squared Error (RMSE) $= \sqrt{\frac{1}{|M|}\sum_{i \in M} (\tilde{D}_i - D_i)^2}$. 
%All metrics are averaged across the test set. 

%All metrics are computed over pixels in $M$ and averaged across the test set. 

\begin{figure}[b]
\centering
\includegraphics[width=0.77\linewidth]{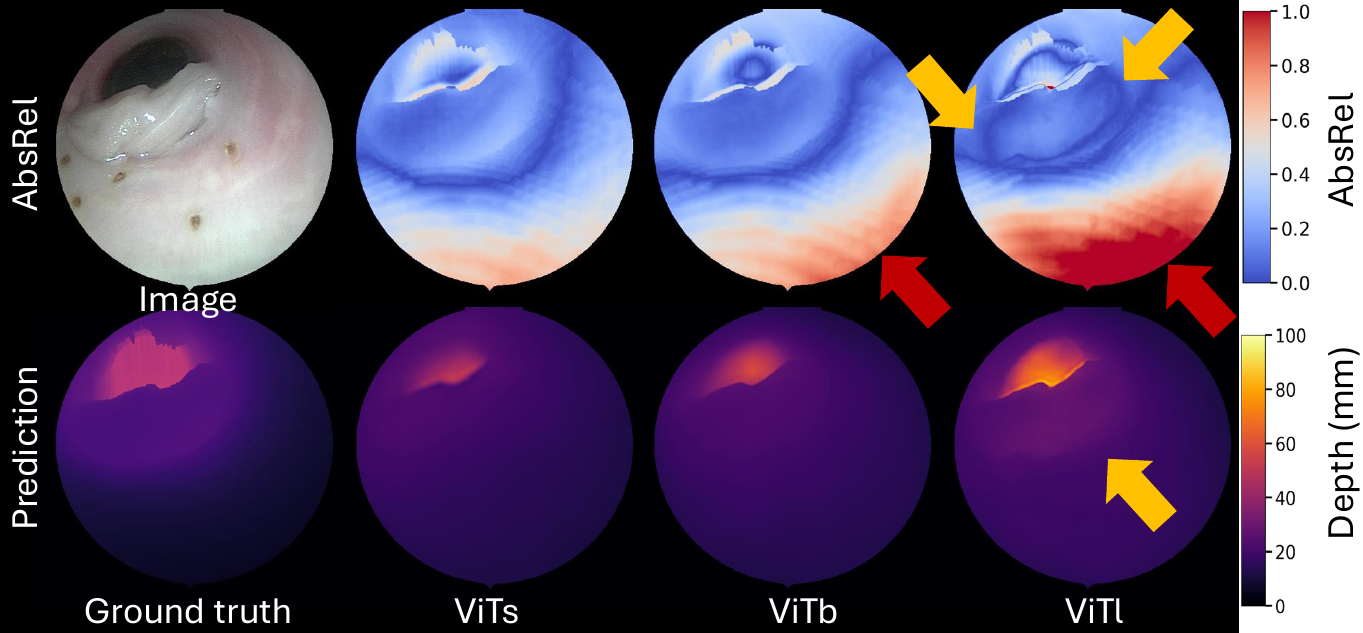}
\caption{
Qualitative results with different size backbones (ViTs, ViTb, and ViTl). The top row shows the AbsRel error maps whereas the bottom row shows the predicted depth maps. EndoOmni pretrained weights were used for each model. Larger backbones  capture boundary depth information more accurately (yellow arrows). This is consistent with Fig.~\ref{backbone}(c–d) where  larger backbones have  lower RMSE. However, larger errors (red arrows) are still observed in homogeneous regions. }
\label{qualitative}
\end{figure}

\noindent \textbf{\underline{Depth estimation performance.}} Table~\ref{main_table} summarizes the performance of the compared models. UDA is a widely used approach that provides a baseline performance for MDE, and using pretrained weights leads to improvement. RG-UDA only improves $\delta_1$, which suggests that appearance alignment alone is insufficient and the domain gap persists. DFA performs worse due to unstable adversarial optimization of separate encoders, which lack shared semantic information from synthetic domain. In contrast, the ablation study (gray) shows that adding domain-invariant feature learning (UDA + DIFL) outperforms both UDA and DFA. This highlights the benefit of aligning latent feature distributions using a shared encoder, which allows the network to leverage context from both domains. Latent consistency regularization alone (UDA + CON) slightly reduces performance, as it may lead the alignment toward incorrect latent directions. However, 
% our approach combining DIFL and CON (UDA + DIFL + CON)
% % distribution-level alignment and directional consistency in the latent space
% achieves the best performance across all metrics \xx{for both masked regions for both overall and task-specific performance.}.
our approach combining DIFL and CON (UDA + DIFL + CON)
achieves the best performance  for either mask ROI.
This suggests that 
distribution alignment reduces global domain gap by learning domain-invariant features, and directional consistency further aligns corresponding features across domains. These help the shared encoder effectively leverage both  domains. %Finally, we further validate the performance by using different pretrained weights.

\noindent \textbf{\underline{Backbone and pretrained weights analysis.}} Fig.~\ref{backbone}(c-d)  show RMSE results for different ViT backbones within different masked regions. The baseline performance tends to drop in both mask regions when using larger size models,  because they may overfit to domain-specific patterns such as texture and lighting, which do not transfer well to the target domain $\mathcal{R}$. However, our method consistently reduces RMSE compared to the baseline across all settings, for either set of pretrained weights. Notably, the largest improvement is observed when using ViTl with EndoOmni pretrained weights in the boundary mask region. This backbone has sufficient capacity to capture fine structural details, and our method further improves its performance by aligning transferable edge features that are preserved across domains, as illustrated in the qualitative results in Fig.~\ref{qualitative}.

\noindent \textbf{\underline{Conclusion.}} In this paper, we present a domain adaptation method for monocular absolute depth estimation in endoscopy by aligning latent features across real and translated synthetic images. Through domain-invariant feature learning and latent consistency for aligning latent features, our approach outperforms existing methods on both absolute and relative metrics. Future work will explore application to other endoscopic procedures.

%% file: table/main_table.tex
\begin{table}[t]
\scriptsize
\centering
\caption{Depth estimation performance.
%ViTs is used as the backbone. 
Bold indicates the best performance.%, and $\uparrow/\downarrow$ denote whether higher/lower values are better for each metric. 
The bottom section (gray) shows the ablation study, where the proposed method can be written as Baseline (UDA) + DIFL + CON. }%The two types of  masks can be viewed in Fig.~\ref{backbone}(a-b).}

\begin{tabular}{l|c|ccc|ccc}
\toprule
\multirow{2}{*}{Methods} & \multirow{2}{*}{Weights} 
& \multicolumn{3}{c|}{\textit{Mask}} 
& \multicolumn{3}{c}{\textit{Boundary mask}} \\
\cline{3-8}
& & $\delta_1\uparrow$ & AbsRel$\downarrow$ & RMSE ($\mathrm{mm})\downarrow$ 
  & $\delta_1\uparrow$ & AbsRel$\downarrow$ & RMSE ($\mathrm{mm})\downarrow$ \\
\midrule
Zero-shot & Depth Anything v2  (Metric) \cite{yang2024depth}  & 0.040  & 0.659 & 14.344 &     0.000  &    0.773    &  24.867     \\
% Zero-shot & endo & 0.040  & 0.659 & 14.344 &     0.000  &    0.773    &  24.867     \\
\midrule
Baseline (UDA)     & - & 0.294  & 0.559 & 10.614 &     0.179  &    0.456    &  16.000     \\
Baseline (UDA) \cite{widya2021self}     & Depth Anything v2  (Metric) \cite{yang2024depth} & 0.365  & 0.264 & 6.252 &     0.144  &    0.326    &  11.379     \\
RG-UDA \cite{rau2023task}   & Depth Anything v2  (Metric) \cite{yang2024depth} &   0.405     &   0.279    &  6.978      &   0.095    &    0.380    &   13.050    \\
DFA \cite{karaoglu2021adversarial}     & Depth Anything v2  (Metric) \cite{yang2024depth} &    0.322    &    0.542   &   8.683    &      0.348  &   0.294      &    10.862   \\
\midrule

\rowcolor{gray!10}
Baseline + DIFL          & Depth Anything v2  (Metric) \cite{yang2024depth}            & 0.460 & 0.284 & 5.937 & 0.251 & 0.283 & 10.238 \\
\rowcolor{gray!10}
Baseline + CON          & Depth Anything v2  (Metric) \cite{yang2024depth}           & 0.359 & 0.288 & 6.389 & 0.376 & 0.276 & 9.850 \\
\rowcolor{gray!10}
\textbf{Proposed} & Depth Anything v2  (Metric) \cite{yang2024depth} &  0.567 &  0.254 & 4.981  & \textbf{0.487}  &  \textbf{0.221} & \textbf{8.236}  \\
\rowcolor{gray!10}
\textbf{Proposed} & EndoOmni \cite{tian2024endoomni}  & \textbf{0.596}  & \textbf{0.238}  & \textbf{4.859}  & 0.431  &  0.236 & 8.600  \\
\bottomrule
\end{tabular}
\label{main_table}
\end{table}